\documentclass[letterpaper]{article} 
\usepackage{aaai2026}  
\usepackage{times}  
\usepackage{helvet}  
\usepackage{courier}  
\usepackage[hyphens]{url}  
\usepackage{graphicx} 
\usepackage{natbib}  
\usepackage{caption} 
\usepackage{amsmath}   
\usepackage{amssymb}   
\usepackage{bm}        
\usepackage{booktabs}   
\usepackage{multirow}    
\usepackage{algorithm}
\usepackage{algorithmic}
\usepackage[colorlinks, linkcolor=blue, urlcolor=blue, citecolor=black]{hyperref}
\usepackage{newfloat}
\usepackage{listings}
\DeclareCaptionStyle{ruled}{labelfont=normalfont,labelsep=colon,strut=off} 
\floatstyle{ruled}
\newfloat{listing}{tb}{lst}{}
\floatname{listing}{Listing}
\title{Hierarchical Direction Perception via Atomic Dot-Product Operators for Rotation-Invariant Point Clouds Learning}
\author{
	Chenyu Hu\textsuperscript{\rm 1},
	Xiaotong Li\textsuperscript{\rm 1}\thanks{Corresponding author.},
	Hao Zhu\textsuperscript{\rm 1},
	Biao Hou\textsuperscript{\rm 1}
}
\affiliations {
	\textsuperscript{\rm 1}Xidian University, Xi'an, China\\
	hcy299792@stu.xidian.edu.cn, 
	\{lixiaotong, haozhu\}@xidian.edu.cn, 
	avcodec@163.com
}

\usepackage{bibentry}

\begin{document}
\pagestyle{plain}
\maketitle
\thispagestyle{plain}

\begin{abstract}
	Point cloud processing has become a cornerstone technology in many 3D vision tasks. However, arbitrary rotations introduce variations in point cloud orientations, posing a long-standing challenge for effective representation learning. The core of this issue is the disruption of the point cloud's intrinsic directional characteristics caused by rotational perturbations. Recent methods attempt to implicitly model rotational equivariance and invariance, preserving directional information and propagating it into deep semantic spaces. Yet, they often fall short of fully exploiting the multiscale directional nature of point clouds to enhance feature representations. To address this, we propose the Direction-Perceptive Vector Network (DiPVNet). At its core is an atomic dot-product operator that simultaneously encodes directional selectivity and rotation invariance—endowing the network with both rotational symmetry modeling and adaptive directional perception. At the local level, we introduce a Learnable Local Dot-Product (L2DP) Operator, which enables interactions between a center point and its neighbors to adaptively capture the non-uniform local structures of point clouds. At the global level, we leverage generalized harmonic analysis to prove that the dot-product between point clouds and spherical sampling vectors is equivalent to a direction-aware spherical Fourier transform (DASFT). This leads to the construction of a global directional response spectrum for modeling holistic directional structures. We rigorously prove the rotation invariance of both operators. Extensive experiments on challenging scenarios involving noise and large-angle rotations demonstrate that DiPVNet achieves state-of-the-art performance on point cloud classification and segmentation tasks. Our code is available at \url{https://github.com/wxszreal0/DiPVNet}.
\end{abstract}


\begin{figure}[t]
	\centering
	\includegraphics[width=1.0\linewidth]{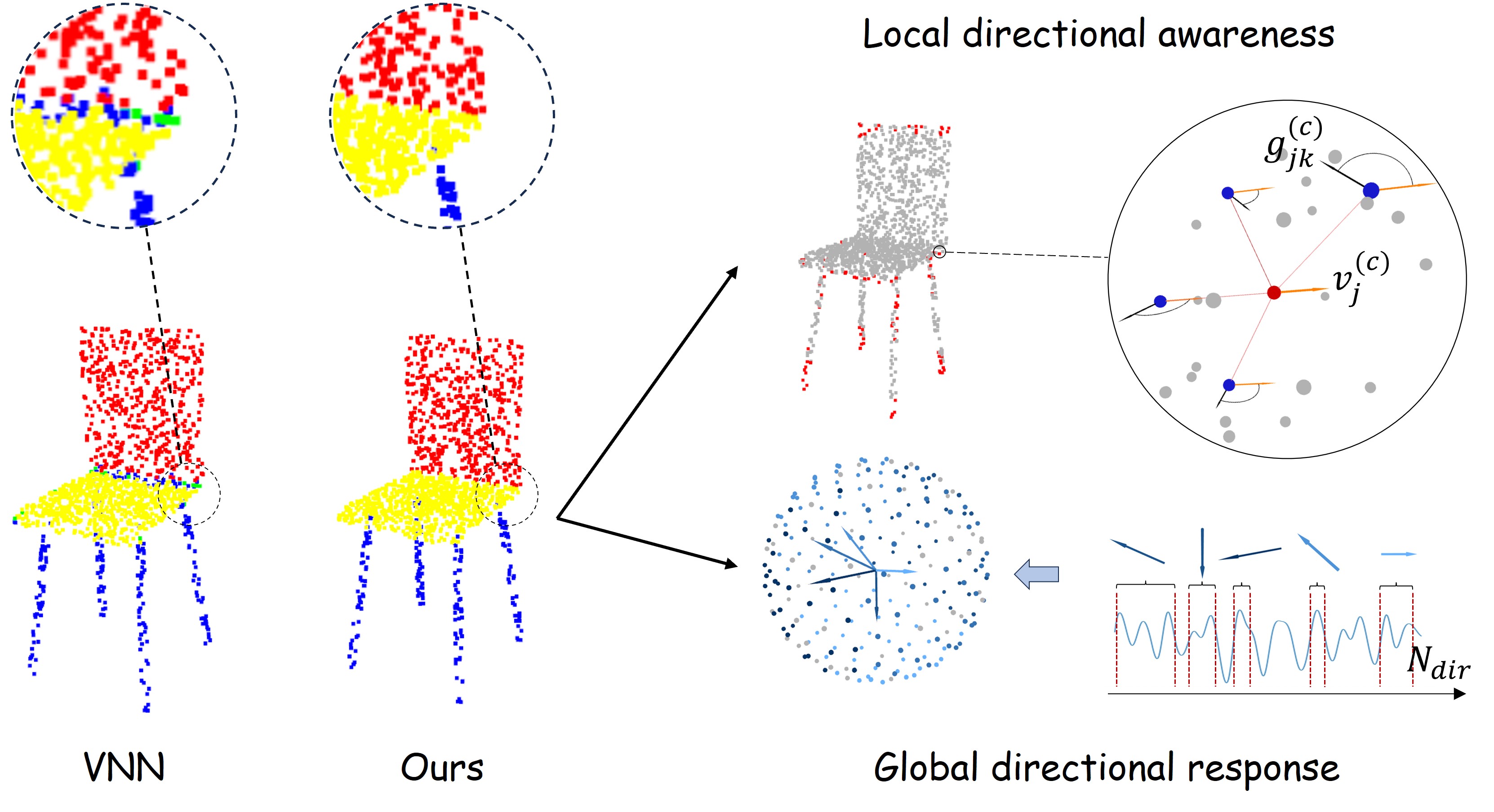}  
	\caption{
		We extract direction-sensitive local features via the L2DP operator; concurrently, a global directional response spectrum is constructed through the DASFT module, capturing the directional characteristics of the overall structure.
	}
	\label{fig:dipvnet}
\end{figure}

\section{Introduction}

The widespread adoption of 3D sensing technologies has made point cloud processing a critical component in numerous 3D vision applications \cite{guo2020deep,feng2023clustering,jia2025ai,feng2024lsk3dnet}, including autonomous driving scene understanding and embodied AI interaction. However, conventional point cloud representation learning methods face a fundamental challenge in real-world deployments \cite{dym2020universality}: arbitrary rotations in 3D space can alter the spatial distribution of point clouds. As a result, traditional networks \cite{qi2017pointnet, qi2017pointnet++, xiang2021walk} may map different orientations of the same object to inconsistent feature representations, ultimately degrading the performance of downstream tasks under rotational disturbances.

At the root of this issue lies the directional nature of point cloud features across multiple scales. On the local scale, discriminative cues are embedded in features such as edge orientation and surface normals; on the global scale, high-level geometric priors arise from principal axis directions, inter-part angles, and overall structural symmetries. These intrinsic directional characteristics are disrupted under arbitrary rotations, implying that robust point cloud representations must explicitly model and leverage such multiscale directional information to achieve rotation resilience.

To tackle this challenge, earlier methods focused on explicit modeling by independently extracting features along predefined directions—such as dividing local neighborhoods into directional sectors \cite{jiang2018pointsift} or introducing orientation density functions (ODF) \cite{sahin2022odfnet}—thus encoding directional information in a manually structured manner. While this mitigated orientation-related inconsistencies to some extent, more recent implicit modeling paradigms  incorporate the rotational symmetry of 3D space directly into the network design. These methods produce outputs that are either equivariant (rotate synchronously with the input)\cite{thomas2018tensor,schutt2021equivariant, poulenard2021functional} or invariant (remain unchanged) \cite{zhang2019rotation,chou20213d,gu2021learning} under input rotations, thereby allowing directional information to be implicitly propagated to deeper semantic layers.

Although both paradigms have demonstrated promising results under rotational perturbations, they remain unsatisfactory in critical ways. Explicit approaches often rely on fixed directional partitions or handcrafted statistical schemes, lacking the adaptability to learn from non-uniform local point distributions, which limits the robustness and discriminability of their directional representations. Implicit approaches, while successfully preserving directional information through rotation-equivariant or invariant formulations, often fail to exploit this information efficiently at the feature representation level. For instance, in VNN \cite{deng2021vector}, the non-linear layers are guided by a single learned global direction vector, which is used to gate vector neurons through inner products. This reliance on a single global direction fails to capture the complex hierarchical directional structure inherent in real-world point clouds.

To enable networks to preserve rotational symmetry while adaptively perceiving multiscale directional features, we propose DiPVNet, a framework constructed upon atomic dot-product operators. Our key insight lies in revealing the intrinsic property of the dot-product operation: it acts as a directional filter that simultaneously exhibits directional selectivity and rotation invariance. Based on this, DiPVNet avoids the need to explicitly model predefined directional features. Instead, as visualized in Figure 1, the directional selectivity of dot-product operators enables adaptive perception of local directions, while the hierarchical aggregation of global directional responses builds discriminative multiscale representations. The rotation invariance of the operator ensures robustness to arbitrary rotational transformations.

Specifically, our contributions are as follows:
\begin{itemize}
	\item We identify and exploit the dual property of the dot-product operation—directional selectivity and rotation invariance—to design atomic dot-product operators that empower the network to perceive directional structures adaptively while maintaining rotational symmetry.
	\item We propose a Learnable Local Dot-Product (L2DP) operator, which performs differentiable dot-products between a center point and its neighbors, enabling adaptive learning of local directional features and improving the network’s ability to handle non-uniform local structures.
	\item We propose the Direction-Aware Spherical Fourier Transform (DASFT). It uses dot-products to project point clouds onto spherical vectors, yielding a global directional response spectrum. This captures global directional features while reducing geometry misinterpretation risks from over-reliance on local features.
\end{itemize}

Extensive experiments under various challenging scenarios—including noise, large-angle rotations, and occlusions—demonstrate that DiPVNet outperforms existing rotation-robust methods across multiple benchmarks. Our method accurately identifies both discriminative global directional structures and critical local regions, significantly improving the robustness and accuracy of point cloud classification, segmentation, and other downstream 3D tasks.

\section{Related Work}
To address the challenge of rotational perturbations in 3D point cloud analysis, existing research has explored various strategies to enhance rotational robustness. These approaches can be broadly categorized into two main directions: explicit directional encoding and implicit modeling of rotational symmetry.

\subsection{Explicit directional encoding}
Explicit directional encoding focuses on designing specific mechanisms to directly capture directional information in point clouds. For example, common strategies \cite{xia2021soe, wang2023pasiftnet} such as spatial orientation partitioning divide local space into fixed directional quadrants to extract orientation-aware features. ODFNet \cite{sahin2022odfnet} models directionality by partitioning the spherical neighborhood into predefined directional cones and statistically analyzing the point distributions. Although these methods alleviate representation inconsistency caused by orientation variations through explicit directional modeling, their fixed spatial partitioning schemes are limited in adapting to the non-uniform distribution of point clouds.

\subsection{Implicit modeling of rotational symmetry}
Implicit modeling of rotational symmetry aims to construct network architectures that are either equivariant or invariant to 3D rotations, so that the output features co-transform with (equivariance) or remain unaffected by (invariance) input rotations, thereby preserving directional information in deeper semantic spaces.

\begin{figure*}[t]
	\centering
	\includegraphics[width=1.0\linewidth]{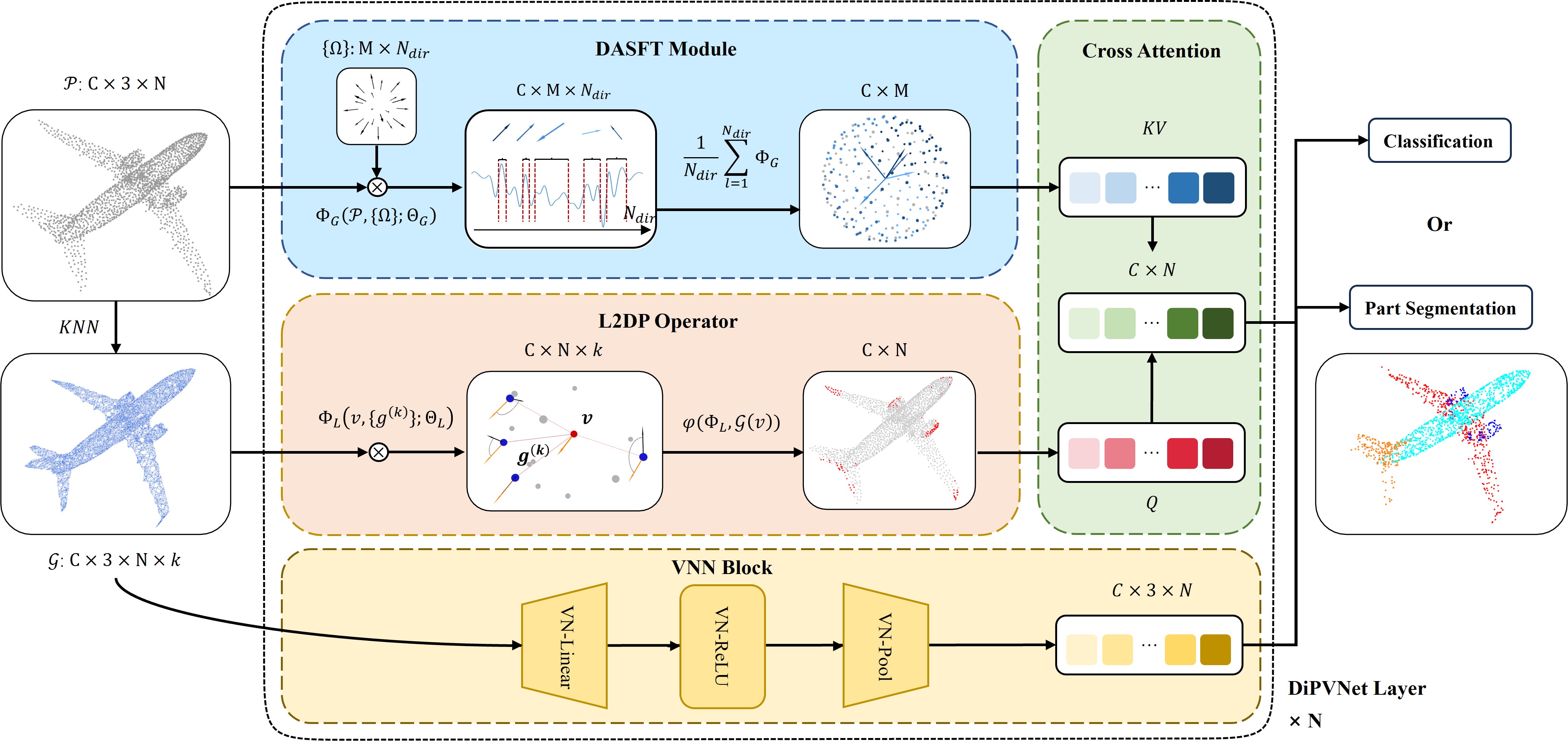}  
	\caption{
		DiPVNet single-layer architecture. Point cloud features $\mathcal{P}$ are transformed into graph features $\mathcal{G}$ via KNN graph construction. In the DiPVNet layer, the VNN Block models rotation equivariance, while concurrently the L2DP operator processes graph features through $\Phi_L$ and extracts local directional features via aggregation mapping $\varphi(\cdot,G(v))$, and the DASFT module constructs a global directional response spectrum through the dot-product operator $\Phi_G$ between point clouds and spherical sampling vectors. Local and global features are fused via cross-attention mechanism, with the output discriminative directional features utilized for downstream tasks.
	}
	\label{fig:drinet}
\end{figure*}

\subsubsection{Invariant Representation Learning}
Earlier works relied on handcrafted descriptors \cite{chen2022devil,gu2021learning,gu2022enhanced,li2021rotation,zhang2024risurconv} or applied PCA-based \cite{li2021closer,yu2020deep,xiao2019endowing} alignment to eliminate orientation discrepancies. Modern approaches \cite{zhang2020learning,kim2020rotation,zhang2024risurconv} often learn Local Reference Frames (LRFs) and model features within the predicted local coordinate systems to achieve invariance. However, their reliance on local features may lead to misinterpretations of the global geometric structure. Notably, the dot-product operator, due to its inherent rotational invariance, has been widely adopted to construct stable invariant information flows. For instance, SGMNet \cite{xu2021sgmnet} uses dot-product operations to build local rotation-invariant features. However, its sorting mechanism disrupts the original spatial orientation relationships between point pairs, and it lacks a feature aggregation strategy to further enhance directional awareness.

\subsubsection{Equivariant Representation Learning}
Tensor Field Networks (TFN)\cite{thomas2018tensor} leverage spherical harmonics to construct group-equivariant convolutional kernels, theoretically enabling the handling of arbitrary rotations. Nonetheless, the high-dimensional harmonic expansions introduce considerable computational overhead. Inspired by TFNs, Vector Neuron Networks (VNNs)\cite{deng2021vector} extend neural outputs from scalars to 3D vectors and employ vectorized operations to impose equivariant constraints implicitly at the network level. Subsequent developments \cite{satorras2021n, lin2023lie,melnyk2024tetrasphere} have further strengthened the stable propagation of directional information into deeper semantic layers. However, these methods \cite{luo2022equivariant, su2022svnet, jing2020learning} still struggle to efficiently exploit directional cues at the feature representation level. For example, VNN selects a single global direction vector to filter vector neurons, which fails to capture the intrinsic, fine-grained directional diversity inherent in point clouds. Other VNN-based works, such as VN-Transformer \cite{assaad2022vn}, primarily utilize the dot-product to construct similarity matrices rather than to explicitly model directional information.

Recent studies have focused on joint modeling of rotational equivariance and invariance, typically through dual-branch architectures \cite{chen2024local} or feature decoupling mechanisms \cite{zhang2023parot}. These designs preserve directional information via the equivariant branch and extract rotation-robust features through the invariant branch, achieving stable representations under arbitrary rotations. However, current methods still face certain limitations: directional information is often indirectly recovered based on predefined geometric constraints, while the global distribution of orientations remains under-explored and under-modeled.
\section{Method}
In this section, we first present the Atomic Dot-Product Operator. Building on this, we propose DiPVNet for direction-aware point cloud representation. Further, we introduce two key components that leverage the operator to enhance multi-scale directional perception: the L2DP operator and the DASFT module.

\subsection{Atomic Dot-Product Operator}
We encapsulate the dot-product operation into a differentiable atomic operator:
\begin{equation}
	\Phi \left( \mathbf{a}, \{\mathbf{b}_i\}; \boldsymbol{\Theta} \right)
	= \mathrm{FFN} \left( \{ \langle \mathbf{a} \cdot \mathbf{b}_i \rangle \}_{i=1}^K; \boldsymbol{\Theta} \right)\,,
\end{equation}
where $\{\mathbf{b}_i\}_{i=1}^K$ denotes a set of associated vectors, and $\boldsymbol{\Theta}$ represents the learnable parameters of the FFN (Feed-Forward Network). This operator inherently combines directional awareness with rotational robustness, serving as the fundamental computational unit and core component for constructing DiPVNet.

\subsection{DiPVNet}
Building upon the atomic dot-product operator, we propose DiPVNet—a point cloud representation learning framework centered around dot-product operators. Figure 2 illustrates the DiPVNet architecture and its module designs. This framework hierarchically models cross-scale directional characteristics of point clouds while preserving rotational symmetry.

First, we propose the L2DP operator, which extends the atomic dot-product operator for local feature extraction:
\begin{equation}
	\Phi_{\mathrm{L}} \left( \mathbf{v}, \{\mathbf{g}^{(k)}\}; \boldsymbol{\Theta} \right) 
	= \mathrm{FFN}_{\mathrm{L}} \left( \langle \mathbf{v} \cdot \{\mathbf{g}^{(k)}\} \rangle; \boldsymbol{\Theta}_{\mathrm{L}} \right)\,,
\end{equation}
where $\{\mathbf{g}^{(k)}\} \in \mathcal{G}(\mathbf{v})$ denotes the set of $k$-nearest neighbors of the central point $\mathbf{v}$. 

Finally, adaptive integration of directional information within the neighborhood is achieved through an invariant feature aggregation mapping $\phi(\cdot, \mathcal{G}(\mathbf{v}))$, designed with two implementations: Direct Linear Projection (DLP) and Statistic-Aware Projection (SAP), each tailored for distinct application scenarios:
\begin{equation}
	f_{\mathrm{L2DP}} (\mathbf{v}) 
	= \phi \left[ \Phi_{\mathrm{L}} \left( \mathbf{v}, \{\mathbf{g}^{(k)}\}; \boldsymbol{\Theta}_{\mathrm{L}} \right), \mathcal{G}(\mathbf{v}) \right]\,.
\end{equation}

This operator can adaptively perceive and learn local directional features, significantly enhancing the model's representation capability for non-uniform local structures.

Second, to model the global directional characteristics of point clouds, we compute the dot-product between the raw point cloud $\mathcal{P}$ and a set of sampled vectors $\{\boldsymbol{\Omega}_l\}_{l=1}^{N_{\mathrm{dir}}}$ on the sphere $S^2$. This operation is equivalent to the Fourier transform of the point cloud in the spherical frequency domain $\mathcal{F}(\mathcal{P}, \{\boldsymbol{\Omega}_l\})$. We thus define the Global Dot-Product Operator:
\begin{equation}
	\Phi_{\mathrm{G}} \left( \mathcal{P}, \{\boldsymbol{\Omega}_l\}; \boldsymbol{\Theta} \right) 
	= \mathrm{FFN}_{\mathrm{G}} \left( E(\mathcal{P}, \{\boldsymbol{\Omega}_l\}); \boldsymbol{\Theta}_{\mathrm{G}} \right)\,,
\end{equation}
where
\begin{equation}
	E(\mathcal{P}, \{\boldsymbol{\Omega}_l\}) 
	= \left| \mathcal{F}(\mathcal{P}, \{\boldsymbol{\Omega}_l\}) \right|^2
\end{equation}
denotes the response power spectrum of point cloud $P$ along the sampled directions $\{\boldsymbol{\Omega}_l\}_{l=1}^{N_{\mathrm{dir}}}$. 

The rotation-invariant global descriptor is then obtained through spherical averaging:
\begin{equation}
	f_{\mathrm{DASFT}} (\mathcal{P}) 
	= \frac{1}{N_{\mathrm{dir}}} \sum_{l=1}^{N_{\mathrm{dir}}} 
	\Phi_{\mathrm{G}} \left( \mathcal{P}, \boldsymbol{\Omega}_l; \boldsymbol{\Theta}_{\mathrm{G}} \right)\,,
\end{equation}
where $N_{\mathrm{dir}}$ is the number of spherical sampling directions. 

This entire pipeline constitutes the DASFT module. By learning multiscale directional response spectra, it effectively models global directional characteristics of point clouds, mitigating potential misjudgments of overall geometric structures that may arise from relying solely on local features.

Finally, a cross-attention mechanism fuses the invariant features: L2DP serves as the query, while DASFT provides the key/value to adaptively guide local features. Concurrently, the equivariant features from the baseline VNN Block are projected onto a learned canonical basis to produce rotation-robust scalar tokens, which are then concatenated with the fused invariant features for downstream tasks.

\subsection{Learnable Local Dot-Product (L2DP)}
\subsubsection*{Rotation-Invariant Local Representation}
Building upon the proven rotational invariance of the dot-product (proof in Appendix 1.1), we define the rotation-invariant local geometric representation for each point $v_j$ as:
\begin{equation}
	I_j^{(\mathcal{N})} = \{ \langle v_j,\, v_k \rangle \mid v_k \in \mathcal{N}_j \}\,,
\end{equation}
where \(\mathcal{N}_j\) denotes the neighborhood set of point \(v_j\) comprising its \(K\) nearest neighbors. If two point clouds, \(\mathcal{P}\) and \(\mathcal{P}'\), are equivalent under the \(\mathrm{SO}(3)\) group (i.e., \(\exists\, R \in \mathrm{SO}(3)\) such that \(\mathcal{P}' = R \cdot \mathcal{P}\)), then their set of local invariants satisfies:
\begin{equation}
	I_j^{(\mathcal{N})} = I_j^{(\mathcal{N})'} \quad \forall j\,.
\end{equation}

This means that the local invariant set \(I_j\) fully characterizes the equivalence class of the point cloud under the action of the rotation group.
\begin{figure}[t]
	\centering
	\includegraphics[width=1.0\linewidth]{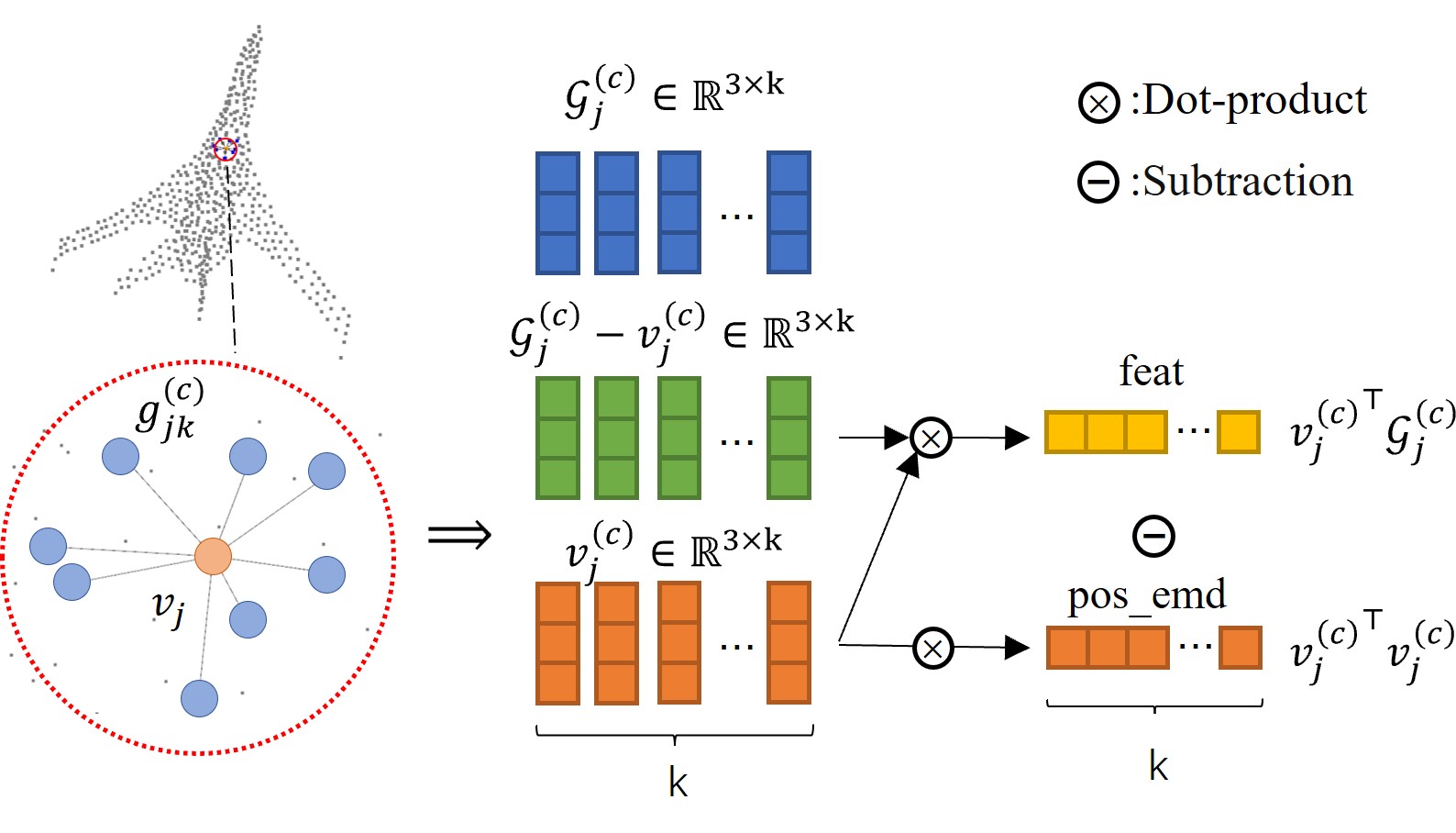}  
	\caption{
		In the partial process of the L2DP operator acting on the $j$-th center point $v_j$, the center point feature is replicated $k$ times and subtracted from neighboring features in the $k$-nearest neighbor neighborhood $\mathcal{G}_j$; a dot-product operation is performed to obtain directional information relative to the center point and its positional encoding; and the result is fed into FFN.
	}
	\label{fig:drinet}
\end{figure}
However, the original definition \(I_j^{(\mathcal{N})}\) only describes the degree of directional correlation between the center point \(v_j\) and its neighbor point \(v_k\), but loses the characteristics of the center point itself. In order to enhance the geometric perception, the relative position coding is further introduced:
\begin{equation}
	I_j^{(\mathcal{N},\mathrm{rel})} = \{ \langle v_j,\, v_k - v_j \rangle \mid v_k \in \mathcal{N}_j \}\,.
\end{equation}

The ``rel'' tag indicates the introduction of relative position encoding.

In practice, for each point \(v_j\) in the point cloud 
\(
\mathcal{P} = \{ v_j \in \mathbb{R}^{c \times 3} \}_{j=1}^{N},
\)
we can obtain its \(K\) nearest neighbor features to form the local group
\(
\mathcal{G}_j = \{ g_j^{(k)} \in \mathbb{R}^{c \times 3} \}_{k=1}^{K}.
\)
The aggregated local neighborhood is then given by
\(
\mathcal{G} = \{ \mathcal{G}_j \}_{j=1}^{N}
\) \cite{yu2024equivariant}. Therefore, we can obtain the local dot-product invariant
\begin{equation}
	I_j^{(\mathcal{G})} = \{ \langle v_j,\, g_{jk} \rangle \mid k = 1, \dots, K \}\,.
\end{equation}

In the L2DP module, we are actually dealing with:
\begin{equation}
	I_j^{(\mathcal{G},\mathrm{rel})} = \{ \langle v_j,\, g_{jk} - v_j \rangle \mid k = 1, \dots, K \}\,.
\end{equation}

This means that the point product result, computed channel-wise over the 3D spatial dimensions, is $\langle v_j, g_{jk} \rangle - \langle v_j, v_j \rangle,$ where $\langle v_j, g_{jk} \rangle$ contains local geometric information, and \(v_j^{(c)\top} v_j^{(c)}\) can be interpreted as injecting position coding \cite{qiu2022spe}, which rectifies the inherent disorder of the point cloud and improves the generalization ability of subsequent attention calculations. The process of obtaining local invariants is shown in Figure~3.
We implement the feature extraction process through the dot-product operator:
\begin{equation}
	\Phi_{\mathrm{L}} \left( I_j^{\mathrm{(G,rel)}}; \boldsymbol{\Theta}_{\mathrm{L}} \right)
	= \mathrm{FFN}_{\mathrm{L}} \left( \left\langle \mathbf{v}_j, \mathbf{g}_j^{(k)} - \mathbf{v}_j \right\rangle; \boldsymbol{\Theta}_{\mathrm{L}} \right)\,,
\end{equation}
thereby adaptively enhancing neighborhood directional awareness while preserving rotation invariance, ultimately strengthening the model's local modeling capability for non-uniform structures.

\subsubsection*{Invariant Feature Aggregation}

To integrate local invariants into point-level features while preserving rotation invariance, we propose two learnable aggregation mappings $\phi(\cdot, \mathcal{G}(\mathbf{v}))$ with complementary characteristics. 

The Direct Linear Projection (DLP) implements $\phi_{\mathrm{DLP}} (\cdot, \mathcal{G}(\mathbf{v})) : \mathbb{R}^{c \times K} \to \mathbb{R}^c$, which directly maps all $K$ dot-product outputs to a high-dimensional semantic space:
\begin{equation}
	f_{\mathrm{DLP}}(\mathbf{v}) 
	= \mathrm{Layernorm} \left( \phi_{\mathrm{DLP}} \left( \Phi_{\mathrm{L}} \left( I^{\mathrm{(G,rel)}}; \boldsymbol{\Theta}_{\mathrm{L}} \right) \right) \right)\,.
\end{equation}

This approach preserves full neighbor interaction information but incurs linearly growing computational complexity with $K$, making it optimal for small-scale neighborhoods.

Conversely, the Statistic-Aware Projection (SAP) employs feature compression through statistical pooling, which computes the maximum, variance, and average of the relative features, denoted as $I_j^{\max}$, $I_j^{\mathrm{var}}$, and $I_j^{\mathrm{avg}}$, respectively. This is followed by $\phi_{\mathrm{SAP}} (\cdot, \mathcal{G}(\mathbf{v})) : \mathbb{R}^{c \times 3} \to \mathbb{R}^c$:
\begin{equation}
	f_{\mathrm{SAP}}(\mathbf{v}) 
	= \mathrm{Dropout} \left( \phi_{\mathrm{SAP}} \left( \left[ I_j^{\max} \Vert I_j^{\mathrm{var}} \Vert I_j^{\mathrm{avg}} \right] \right) \right)\,.
\end{equation}

SAP sacrifices geometric details for enhanced computational efficiency and noise robustness, better suited for large-scale neighborhoods requiring long-range modeling.

\subsubsection*{Operator Formalization}
Finally, we encapsulate the aforementioned process into the L2DP operator.
\begin{equation}
	f_{\mathrm{L2DP}} (\mathbf{v}) 
	= \phi \left[ \Phi_{\mathrm{L}} \left( \mathbf{v}, \{\mathbf{g}^{(k)}\}; \boldsymbol{\Theta}_{\mathrm{L}} \right), \mathcal{G}(\mathbf{v}) \right]\,.
\end{equation}

This operator adaptively perceives and learns local directional characteristics while aggregating features into rotation-invariant point-level representations. The framework significantly enhances the model's capacity for representing non-uniform local structures through its direction-aware learning mechanism.

\begin{figure}[t]
	\centering
	\includegraphics[width=1.0\linewidth]{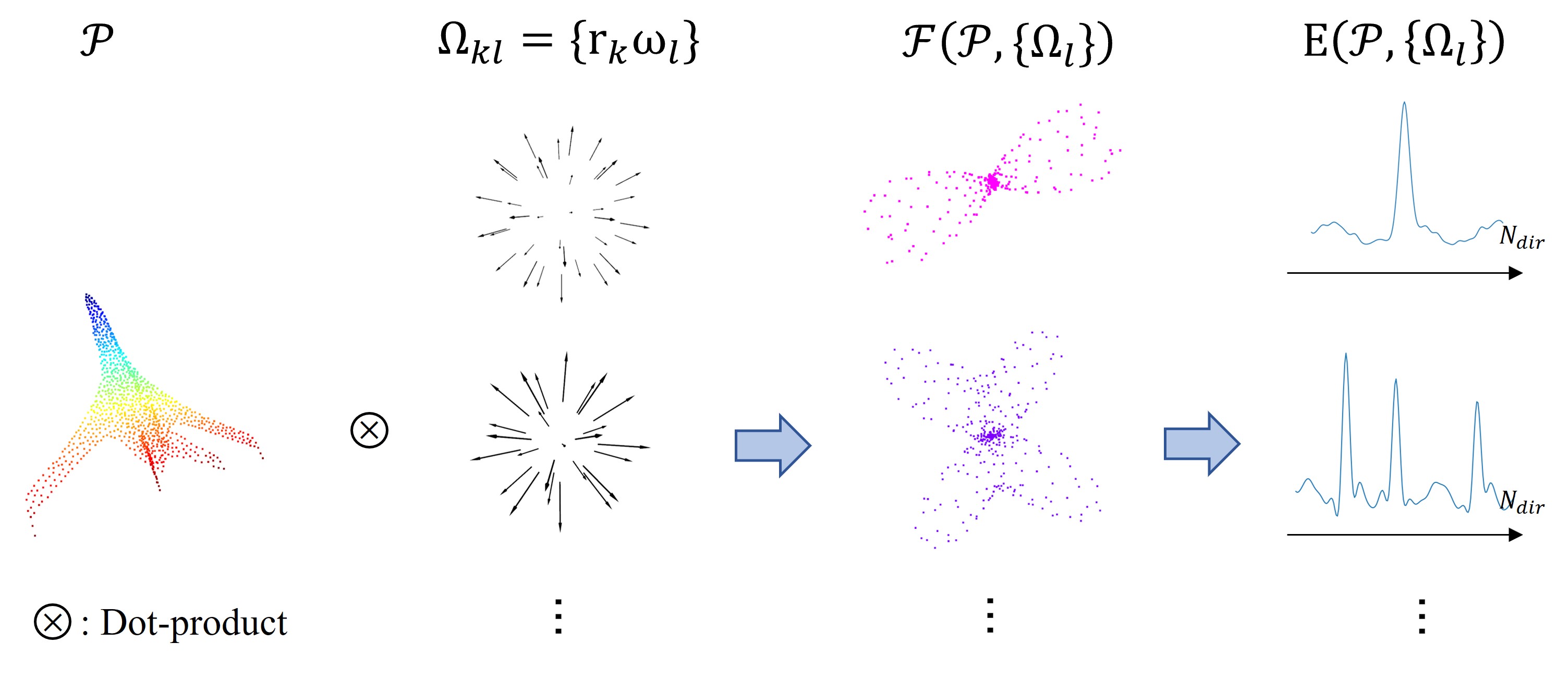}  
	\caption{
		The point cloud $\mathcal{P}$ is projected via dot-products with spherical sampling unit vectors $\Omega$ of varying frequency amplitudes, yielding the spherical frequency-domain response $F(\mathcal{P},\{\Omega\})$ of global features, subsequently constructing the directional response spectrum $E(\mathcal{P},\{\Omega\})$ which characterizes the dominant directions of the point cloud's macroscopic structure.
	}
	\label{fig:drinet}
\end{figure}

\subsection{Direction-Aware Spherical Fourier Transform (DASFT)}
To model the global directional characteristics of point clouds, we can regard the point cloud $\mathcal{P} = \{\mathbf{v}_j \in \mathbb{R}^3\}_{j=1}^N$ as a discrete signal distributed in 3D space. Formally, this can be represented as:
\begin{equation}
	\mathcal{P}(v) = \sum_{j=1}^n \delta(v - v_j), \quad \delta(v) = \begin{cases} +\infty & v = 0 \\0 & \text{otherwise}\end{cases}\,,
\end{equation}
where $\delta(\cdot)$ is the Dirac delta function. This means the point cloud $\mathcal{P}$ can be seen as a superposition of pulsed signals located at $\{ v_j \in \mathbb{R}^{c \times 3} \}_{j=1}^N$.

Under the framework of generalized functions \cite{kondor2007novel}, the Fourier transform of $\mathcal{P}(v)$ is:
\begin{equation}
	\mathcal{F}(\mathcal{P}, \{\boldsymbol{\Omega}\})  = \int \mathcal{P}(v) \exp(-i \Omega^\top v) dv = \sum_{j=1}^n \exp(-i \Omega^\top v_j)\,,
\end{equation}
where $\Omega = r\omega$ is a frequency space vector ($r$: frequency modulus length, $\omega \in S^2$: unit direction vector). The dot-product between $\mathcal{P}$ and $\Omega$ corresponds to the phase term:
\begin{equation}
	\mathcal{F}(\mathcal{P}, \{\boldsymbol{\Omega}\}) = \sum_{j=1}^n \exp(-i r \omega^\top v_j)\,.
\end{equation}

This indicates each point $v_j$ contributes a phase factor $e^{-i r \omega^\top v_j}$ to frequency $r \cdot \omega$, and the spherical spectrum $\mathcal{F}(r \cdot \omega)$ contains complete global information \cite{xu2022unified,son2024intuitive}.
Further, we define the energy spectrum as the modulus square of the Fourier coefficient
\begin{equation}
	E(\mathcal{P}, \{\boldsymbol{\Omega}\}) = \left| \mathcal{F}(\mathcal{P}, \{\boldsymbol{\Omega}\}) \right|^2\,.
\end{equation}

This equation characterizes the structural energy distribution of the point cloud across different frequency scales $r$ and directions $\boldsymbol{\omega}$. Figure 4 visualizes the directional response results, reflecting the multiscale global directional characteristics of the point cloud.

Based on this analysis, we construct the Global Dot-Product Operator:
\begin{equation}
	\Phi_{\mathrm{G}} \left( \mathcal{P}, \{\boldsymbol{\Omega}\}; \boldsymbol{\Theta}_{\mathrm{G}} \right)
	= \mathrm{FFN}_{\mathrm{G}} \left( E(\mathcal{P}, \{\boldsymbol{\Omega}\}); \boldsymbol{\Theta}_{\mathrm{G}} \right)\,,
\end{equation}
where $\Omega = r \cdot \omega$. This operator adaptively fuses energy channels from different directions through learnable channel mapping.

To construct a rotation-invariant descriptor, we uniformly sample $N_{\text{dir}}$ directions $\{\omega_l\}_{l=1}^{N_{\text{dir}}}$ on the unit sphere $S^2$ and perform energy integration:
\begin{equation}
	G(r) = \frac{1}{4\pi}\int_{S^2} E(r \cdot \omega) d\omega \approx \frac{1}{N_{\text{dir}}} \sum_{l=1}^{N_{\text{dir}}} E(r \cdot \omega_l)\,.
\end{equation}

Since $G(r)$ only depends on the modulus $r$, it is rotation-invariant. While its computation (Eq. 21) is a discrete approximation, we empirically verified its robustness to the sampling density $N_{\text{dir}}$ ($\ge 36$) in the Appendix.

Finally, the DASFT module can be formally expressed as:
\begin{equation}
	f_{\mathrm{DASFT}} (\mathcal{P}) 
	= \frac{1}{N_{\mathrm{dir}}} \sum_{l=1}^{N_{\mathrm{dir}}} 
	\Phi_{\mathrm{G}} \left( P, \boldsymbol{\omega}_l; \boldsymbol{\Theta}_{\mathrm{G}} \right)\,.
\end{equation}

Through hierarchical learning of multiscale directional response spectra, this module models the global directional characteristics of point clouds, effectively mitigating potential misjudgments of overall geometric structures that may occur when relying solely on local features \cite{zhang2020global}.

\section{Experiments}
\subsection{Datasets and Tasks}
\subsubsection*{Shape Classification}
We tested the classification performance of our model on the synthetic dataset ModelNet40 \cite{wu20153d} and the real-world dataset ScanObjectNN \cite{uy2019revisiting}. ModelNet40 consists of 40 categories and a total of 12311 CAD mesh models, of which 9843 are used for training and 2468 are used for testing. ScanObjectNN contains 15000 complete objects scanned from 2902 real-world objects, and we use the $\text{OBJ\_BG}$ subset from them.

\subsubsection*{Part Segmentation}
For the segmentation task, we evaluated the segmentation performance of the network on the ShapeNetPart dataset \cite{chang2015shapenet}, which contains 16880 synthetic samples divided into 14006 training samples and 2874 training samples. Including 16 object categories, each category has 2 to 6 sections of annotations, totaling 50 subdivisions.

\subsubsection*{Task Setup}
We follow the general practice of using universal training/testing rotation settings: z/z, z/SO(3), SO(3)/SO(3), Among them, z represents random rotation around the z-axis, and SO(3) represents three-dimensional random rotation at any angle. These rotations are generated and applied to the input point clouds.
\begin{table}[t]
	\centering
	\begin{tabular}{lccc}
		\toprule
		\textbf{Methods} & \textbf{z/z} & \textbf{z/SO(3)} & \textbf{SO(3)/SO(3)} \\
		\midrule
		\multicolumn{4}{c}{\textbf{Rotation-sensitive}} \\ 
		\midrule
		PointNet & 85.9 & 19.6 & 74.7 \\
		DGCNN & \textbf{92.2} & 20.6 & 81.1 \\
		\midrule
		\multicolumn{4}{c}{\textbf{Rotation-robust}} \\ 
		\midrule
		TFN & 89.7 & 89.7 & - \\
		VN-DGCNN & 89.5 & 89.5 & 90.2 \\
		Pose Selector & 90.2 & 90.2 & 90.2 \\
		LGR-Net & 90.9 & 90.9 & \underline{91.1} \\
		OrientedMP & 88.4 & 88.4 & 88.9 \\
		VN-Transformer & - & - & 90.8 \\
		PaRot & 90.9 & \underline{91.0} & 90.8 \\
		TetraSphere & 90.5 & 90.5 & - \\
		Ours & \underline{91.4} & \textbf{91.4} & \textbf{91.4} \\
		\bottomrule
	\end{tabular}
	\caption{The results of each method on ModelNet40 were compared under different training/testing conditions. The overall optimal results are shown in bold and the sub-optimal items are underlined.}
	\label{tab:performance}
\end{table}
\begin{table}[t]
	\centering
	\begin{tabular}{lccc}
		\toprule
		\textbf{Methods} & \textbf{z/z} & \textbf{z/SO(3)} & \textbf{SO(3)/SO(3)} \\
		\midrule
		\multicolumn{4}{c}{\textbf{Rotation-sensitive}} \\ 
		\midrule
		PointCNN & 86.1 & 14.6 & 63.7 \\
		DGCNN & 82.8 & 17.7 & 71.8 \\
		\midrule
		\multicolumn{4}{c}{\textbf{Rotation-robust}} \\ 
		\midrule
		PaRINet + PCA & 83.3 & 83.3 & 83.3 \\
		PCR-cored framework & - & 86.6 & 86.3 \\
		VN-DGCNN & 83.5 & 83.5 & 84.2 \\
		Pose Selector & 84.3 & 84.3 & 84.3 \\
		LGR-Net & 81.2 & 81.2 & 81.4 \\
		TetraSphere & \underline{87.3} & \underline{87.3} & \underline{87.1} \\
		Ours & \textbf{87.5} & \textbf{87.5} & \textbf{87.4} \\
		\bottomrule
	\end{tabular}
	\caption{Under different training/testing conditions, the results of each method on ScanObjectNN are compared. The overall optimal results are shown in bold and the sub-optimal items are underlined.}
	\label{tab:performance}
\end{table}
\subsection{Experimental Setup}
In terms of point cloud sampling, we use farthest point sampling (FPS). In the classification task, we uniformly sample 1024 points, and in the segmentation task, we sample 2048 points. We implemented our model in PyTorch using the official implementation \cite{deng2021vector,li2021closer,melnyk2024tetrasphere}, while using VN-DGCNN as the baseline network and employing the same data augmentation method: random translation within the range of [-0.2,\ 0.2] and random scaling within the range of [2/3,\ 3/2] during the training process. In the classification task, we employ DLP aggregation for ModelNet40 and SAP aggregation for ScanObjectNN; in the segmentation task, we utilize SAP aggregation. When calculating the DASFT, we linearly sample the frequency within the range of [0,12] and select $N_{dir}=36$ in the directional sampling. Consistent with the baseline, we use SGD with an initial learning rate of 0.1 and momentum equal to\ 0.9, as well as a cosine annealing strategy for gradual reduction with a learning rate of 0.001 and minimizing the cross-entropy loss with label smoothing. For the selection of epochs, we trained 200 epochs on ModelNet40 and 250 epochs on ShapeNet, with a batch size of 32.
\subsection{Results}
We categorize existing methods into rotation sensitive and rotation robust classes.
\subsubsection*{Shape Classification}
Tables 1 and 2 present the comparison results between our method and existing methods in classification tasks. The results demonstrate that our method maintains consistent performance across three training/testing sessions. The results under the z/SO(3) setting best reflect the model's generalization performance. For rotation-sensitive methods \cite{li2018pointcnn,wang2019dynamic}, although their performance on z/z\ is decent, they fail to generalize to unknown orientations under the z/SO(3) setting, resulting in a sharp performance decline. Compared with  PCR-cored framework \cite{yu2023rethinking}, Pose Selector \cite{li2021closer} and other rotational equivariant architectures \cite{zhao2022rotation,fuchs2020se}, our method achieves in-depth exploration of multi-scale directional characteristics in point clouds by introducing dot-product operators into each rotational equivariant layer, while simultaneously modeling rotational symmetry.

\subsubsection*{Part Segmentation}
Table 3 shows the comparison results between our method and existing methods in segmentation tasks. We evaluated the model performance using the Mean Intersection over Union (mIoU) on each instance, and the results showed that our model achieved the best performance on the instance. Figure 5 illustrates the segmentation results through visualizations.
\begin{table}[htbp]
	\centering
	\begin{tabular}{lcc}
		\toprule
		\textbf{Methods} & \textbf{z/SO(3)} & \textbf{SO(3)/SO(3)} \\
		\midrule
		\multicolumn{3}{c}{\textbf{Rotation-sensitive}} \\ 
		\midrule
		PointNet & 37.8 & 74.4 \\
		DGCNN & 37.4 & 73.3 \\
		\midrule
		\multicolumn{3}{c}{\textbf{Rotation-robust}} \\ 
		\midrule
		PCR-cored framework & 80.3 & 80.4 \\
		VN-DGCNN & 81.4 & 81.4 \\
		Pose Selector & 81.7 & \underline{81.7} \\
		LGR-Net & 80.0 & 80.1 \\
		OrientedMP & 80.1 & 80.9 \\
		TetraSphere & \underline{82.3} & - \\
		Ours & \textbf{82.5} & \textbf{82.7} \\
		\bottomrule
	\end{tabular}
	\caption{The results of each method are compared on ShapeNetPart dataset under different training/testing conditions, and the evaluation index is the average intersection union ratio of all instances. The overall optimal results are shown in bold and the sub-optimal items are underlined.}
	\label{tab:performance}
\end{table}

\subsection{Ablation Studies}
As shown in the Table 4, Model A incorporates only the DASFT module, and its performance shows no significant improvement over the baseline, indicating that global direction-aware features alone are insufficient to boost performance. Models B and C compare SAP and DLP aggregation strategies under the condition of removing the DASFT module. On the ModelNet40 dataset, which contains relatively low noise interference, SAP aggregation underperforms compared to DLP, confirming the advantage of DLP in preserving local geometric consistency. However, the results of Model C also demonstrate that relying solely on local features can lead to misinterpretation of global structures, resulting in lower performance compared to the optimal model. Model D replaces Cross Attention with a gating mechanism for feature fusion, and the results indicate that static weight allocation fails to achieve the dynamic feature calibration characteristic of attention mechanisms, leading to suboptimal performance.

\begin{figure}[t]
	\centering
	\includegraphics[width=1.0\linewidth]{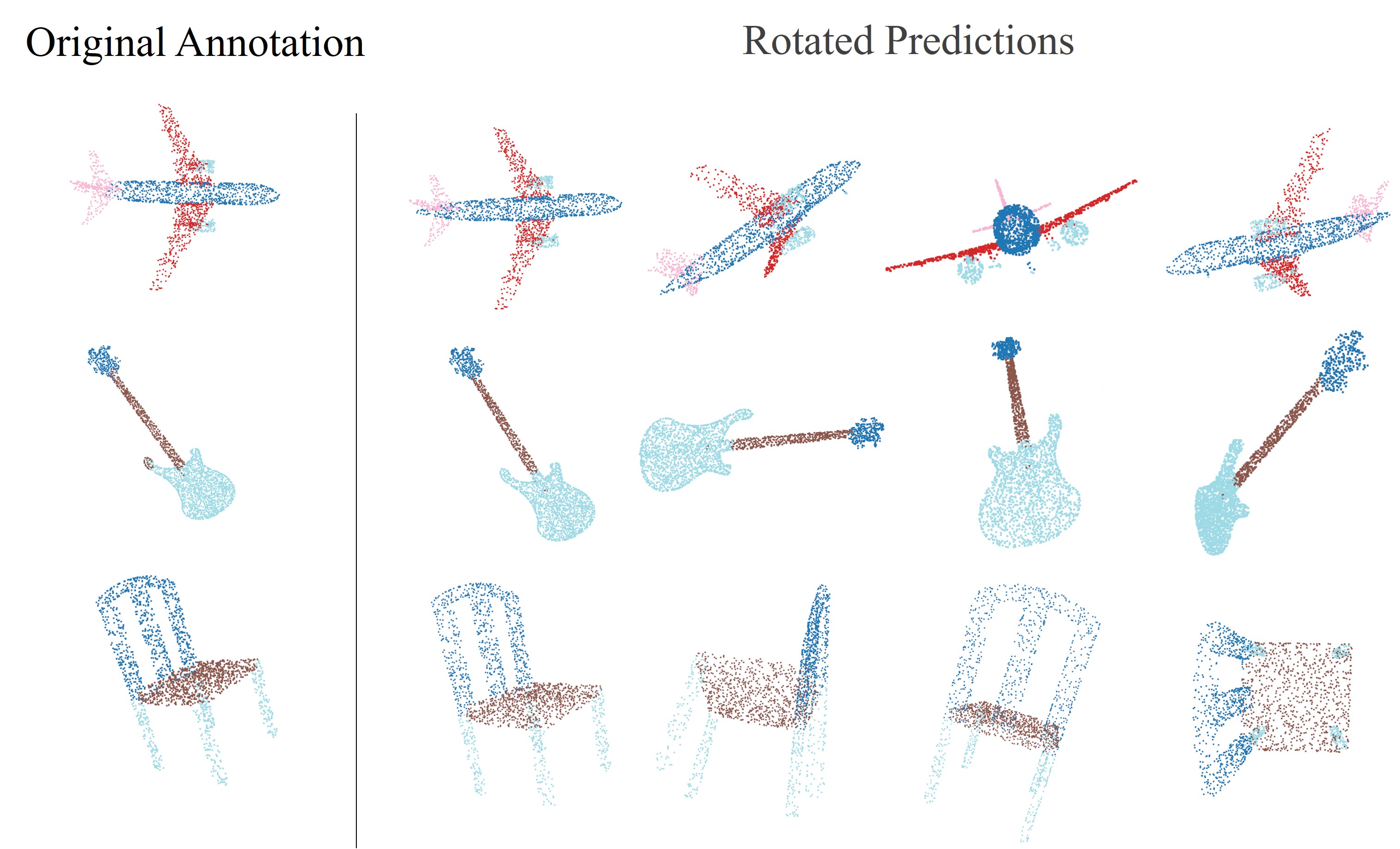}  
	\caption{
		Visualization of segmentation results.
	}
	\label{fig:drinet}
\end{figure}

\begin{table}[t]
	\centering
	\begin{tabular}{@{}c *{2}{c} c *{2}{c} c@{}}
		\toprule
		\multirow{2}{*}{Model} & \multicolumn{2}{c}{L2DP} & \multirow{2}{*}{DASFT} & \multicolumn{2}{c}{Feature Fusion} & \multirow{2}{*}{Acc.} \\
		\cmidrule(lr){2-3} \cmidrule(l){5-6}
		& DLP & SAP & & Gate & CA & \\
		\midrule
		Baseline & \multicolumn{5}{c}{} & 89.5 \\
		A & & & $\checkmark$ & & & 89.5 \\
		B & & $\checkmark$ & & & & 89.9 \\
		C & $\checkmark$ & & & & & 90.6 \\
		D & $\checkmark$ & & $\checkmark$ & $\checkmark$ & & 90.9 \\
		E (Ours) & $\checkmark$ & & $\checkmark$ & & $\checkmark$ & 91.4 \\
		\bottomrule
	\end{tabular}
	\caption{Ablation results assessing component impacts on ModelNet40 under z/SO(3) conditions: L2DP aggregation (DLP or SAP alternatives), DASFT inclusion, and fusion method (CA or Gate).}
	\label{tab:model_comp}
\end{table}

\section{Conclusion}
This paper proposes an efficient direction-aware point cloud rotation-robust framework based on atomic dot-product operators, revealing a new design paradigm that goes beyond module-level combinations from a novel atomic dot-product perspective. The network architecture inspired by this enhances direction-aware capability while modeling rotational symmetry, paving the way for more robust point cloud analysis under arbitrary rotations. Future work will explore the applicability of atomic dot-product operators in other mainstream network architectures, thereby verifying their universal value for performance improvement.

\bibliography{aaai2026}

\clearpage
\appendix

\section{Properties of Dot-Product Operators}
Prior to deriving the Atomic Dot-Product Operator, we first elaborate on the two fundamental properties of the dot-product operation.

\subsection{Rotation Invariance}
Let $V \subset \mathbb{R}^3$ be a three-dimensional Euclidean space and $\mathbf{u}, \mathbf{v} \in V$ be arbitrary vectors. Consider a rotation $R \in \mathrm{SO}(3)$ satisfying $R^\top R=I$. The dot-product $\langle \cdot, \cdot \rangle : V \times V \to \mathbb{R}$ is defined as $\langle \mathbf{u}, \mathbf{v} \rangle = \mathbf{u}^\top \mathbf{v}$. This operation satisfies rotation invariance:
\begin{equation}
	\langle R\mathbf{u}, R\mathbf{v} \rangle 
	= (R\mathbf{u})^\top (R\mathbf{v}) 
	= \mathbf{u}^\top R^\top R \mathbf{v} 
	= \mathbf{u}^\top I \mathbf{v} 
	= \langle \mathbf{u}, \mathbf{v} \rangle.
\end{equation}

This demonstrates the rotation invariance of the dot-product under 3D rotations.

\subsection{Direction Selectivity}
The dot-product fundamentally encodes directional relationships. This directional sensitivity manifests through the natural isomorphism $\phi: V \rightarrow V^\ast$ to the dual space. For any \emph{fixed direction vector} $\mathbf{u} \in V$, define the direction-selective linear functional $\phi_{\mathbf{u}}: V \to \mathbb{R}$ as:
\begin{equation}
	\phi_{\mathbf{u}}(\mathbf{v}) = \langle \mathbf{u}, \mathbf{v} \rangle = \|\mathbf{u}\| \|\mathbf{v}\| \cos \theta_{\mathbf{u},\mathbf{v}}\,,
\end{equation}
where $\theta_{\mathbf{u},\mathbf{v}}$ is the angle between vectors $\mathbf{u}$ and $\mathbf{v}$.
The scalar projection (component) of $\mathbf{v}$ along the direction of $\mathbf{u}$ is then given by:
\begin{equation}
	\operatorname{comp}_{\mathbf{u}}(\mathbf{v}) = \frac{\langle \mathbf{u}, \mathbf{v} \rangle}{|\mathbf{u}|} = \|\mathbf{v}\| \cos \theta_{\mathbf{u},\mathbf{v}}\,.
\end{equation}

This quantifies precisely how much of $\mathbf{v}$ points in the $\mathbf{u}$-direction.
\subsection{Compatibility}
For any rotation $R\in\ SO\left(3\right)$, we have:
\begin{equation}
	\operatorname{comp}_{R\mathbf{u}}(R\mathbf{v}) = \frac{R\mathbf{u} \cdot R\mathbf{v}}{|R\mathbf{u}|}\,.
\end{equation}
Using the norm preservation property:
\begin{equation}
	|R\mathbf{u}| = \sqrt{(R\mathbf{u})^\top R\mathbf{u}} = \sqrt{\mathbf{u}^\top R^\top R \mathbf{u}} = \sqrt{\mathbf{u}^\top \mathbf{u}} = |\mathbf{u}|\,,
\end{equation}
and the rotation invariance of dot-product:
\begin{equation}
	R\mathbf{u} \cdot R\mathbf{v} = \mathbf{u} \cdot \mathbf{v}\,.
\end{equation}
We obtain the compatibility:
\begin{equation}
	\operatorname{comp}{R\mathbf{u}}(R\mathbf{v}) = \operatorname{comp}_{\mathbf{u}}(\mathbf{v})\,.
\end{equation}

This demonstrates the compatibility between directional selectivity and rotation invariance in dot-product operations.

\section{Proof of Rotation Invariance for Each Module}

\subsection{L2DP Operator}
Given a point cloud $\mathcal{P}=\{\{v_j\in\mathbb{R}^{c\times3}\}\}_{i=1}^N$, We can define the 3D rotation group SO(3), it is a Lie group composed of all $3\times3$ real matrices satisfying $R^\top R=I$ and $\det(R)=1$, and its group multiplication is matrix multiplication.\\

Further, the role of the point cloud $\mathcal{P}$, the group element $R\in SO(3)$ is defined as:
\begin{equation}
	R\cdot\mathcal{P} = \{\{Rv_j\}\}_{j=1}^N\,.
\end{equation}
Since rotation does not change the Euclidean distance between points, the original neighborhood $\mathcal{N}_j$ The $k$ points in are still the same batch after rotation, that is, the neighborhood after rotation is $\mathcal{N}_j' = \{v_{j1},\ldots,v_{jk}\}$, therefore, local group $\mathcal{G}_j$ Becomes $\mathcal{G}_j' = \{\{R\cdot g_j^{(k)}\in\mathbb{R}^{c\times3}\}\}_{k=1}^K$ after rotation.\\

For any channel $i=1,\ldots,c$, Record $v_j^{(i)}\in\mathbb{R}^{1\times3}$ as the $i$-th channel of $v_j$, $g_{jk}^{(i)}$ is the $i$-th channel of the $k$-th nearest neighbor of point $v_j$, then the inner product after rotation is:
\begin{equation}
	\langle Rv_j^{(i)}, R(g_{jk}^{(i)}-v_j^{(i)}) \rangle = (v_j^{(i)})^\top R^\top R(g_{jk}^{(i)}-v_j^{(i)})\,.
\end{equation}
Since $R^\top R=I$, it is simplified to
\begin{equation}
	\left(v_j^{(i)}\right)^\top\left(g_{jk}^{(i)}-v_j^{(i)}\right) = \langle v_j^{(i)}, g_{jk}^{(i)}-v_j^{(i)} \rangle\,.
\end{equation}
Therefore, the rotation of the corresponding local inner product of each channel is unchanged. Each channel component of inner product $\langle v_j, g_{jk}-v_j \rangle$ keeps rotation unchanged, i.e
\begin{equation}
	R\cdot I_j^{G,\text{rel}} = I_j^{G,\text{rel}}\,.
\end{equation}
That is, the local point product invariant is strictly rotation invariant.

\subsection{DASFT Module}
Let the rotation matrix be $R\in SO(3)$, and the rotated point cloud be $\mathcal{P}' = \{\{Rv_j\in\mathbb{R}^{c\times3}\}\}_{i=1}^N$. The corresponding generalized function is:
\begin{equation}
	\mathcal{P}'(v) = \sum_{j=1}^n \delta(v-Rv_j)\,.
\end{equation}
The Fourier transform after rotation is:
\begin{equation}
	\mathcal{F}'(\mathcal{P'}, \{\boldsymbol{\Omega}\}) = \sum_{j=1}^n \exp(-i\cdot r\cdot\omega^\top Rv_j)\,,
\end{equation}
where ${\Omega} = r\cdot\omega$. We can construct:
\begin{equation}
	\omega^\top Rv_j = (R^\top\omega)^\top v_j\,.
\end{equation}
Let $\omega'=R^\top\omega\in S^2$, then:
\begin{equation}
	\mathcal{F}'(\mathcal{P'}, \{\boldsymbol{\Omega}\}) = \sum_{j=1}^n \exp(-ir(\omega')^\top v_j) = \mathcal{F}(\mathcal{P}, \{\boldsymbol{\Omega}'\})\,,
\end{equation}
where ${\Omega'} = r\cdot\omega'$. That is, the Fourier coefficient of the rotating point cloud is equivalent to the inverse rotation of the original coefficient in the spherical frequency domain. Since the energy spectrum is defined as $E(\mathcal{P}, \{\boldsymbol{\Omega}\}) = |\mathcal{F}(\mathcal{P}, \{\boldsymbol{\Omega}\})|^2$, so
\begin{equation}
	E'(\mathcal{P}', \{\boldsymbol{\Omega}\}) = |\mathcal{F}'(\mathcal{P}', \{\boldsymbol{\Omega}\})|^2 = |\mathcal{F}(\mathcal{P}, \{\boldsymbol{\Omega}'\})|^2 = E(\mathcal{P}, \{\boldsymbol{\Omega'}\})\,.
\end{equation}
Because $\omega'=R^\top\omega$, the energy spectrum is re parameterized in the frequency direction after rotation, but its mode length remains unchanged.\\

Further, spherical integration is performed on the energy spectrum after rotation and variable replacement is made $\omega'=R^\top\omega$:
\begin{equation}
	G'(r) = \frac{1}{4\pi}\int_{S^2} E'(r\cdot\omega)d\omega = \frac{1}{4\pi}\int_{S^2} E(r\cdot\omega')d\omega\,.
\end{equation}
Since SO(3) rotation keeps the spherical area element unchanged, that is, $d\omega=d\omega'$, the integral becomes
\begin{equation}
	G'(r) = \frac{1}{4\pi}\int_{S^2} E(r\cdot\omega')d\omega' = G(r)\,.
\end{equation}
This means that the power spectrum invariants obtained by spherical area division have strict rotation invariance.

\section{Supplementary for the DASFT Module}
\subsection{Frequency Sampling}
We use linear and logarithmic methods for frequency sampling, that is, given a frequency range $[f_{\min},f_{\max}]$, there are:
\begin{equation}
	r_k=f_{\min}+\frac{k}{m-1}\cdot(f_{\max}-f_{\min}),\quad k=0,1,\ldots,m-1\,,
\end{equation}
\begin{equation}
	\begin{split}
		r_k = \exp\biggl( \ln(f_{\min}+\epsilon) \frac{k}{m-1} \cdot \Bigl[ \ln(f_{\max}+\epsilon)
		-\ln(f_{\min}+\epsilon) \Bigr] \biggr)
	\end{split}\,.
\end{equation}
When it is necessary to retain the main low-frequency structure and sparse the high-frequency details, logarithmic sampling is used. However, in practical tasks, linear sampling is generally used to extract multi-level rotation invariant features, so as to model the multi-scale geometric distribution. Figure 1 shows the appearance of the first five kinds of spectrum when $m=10$.\\
\begin{figure}[t]
	\centering
	\includegraphics[width=1.0\linewidth]{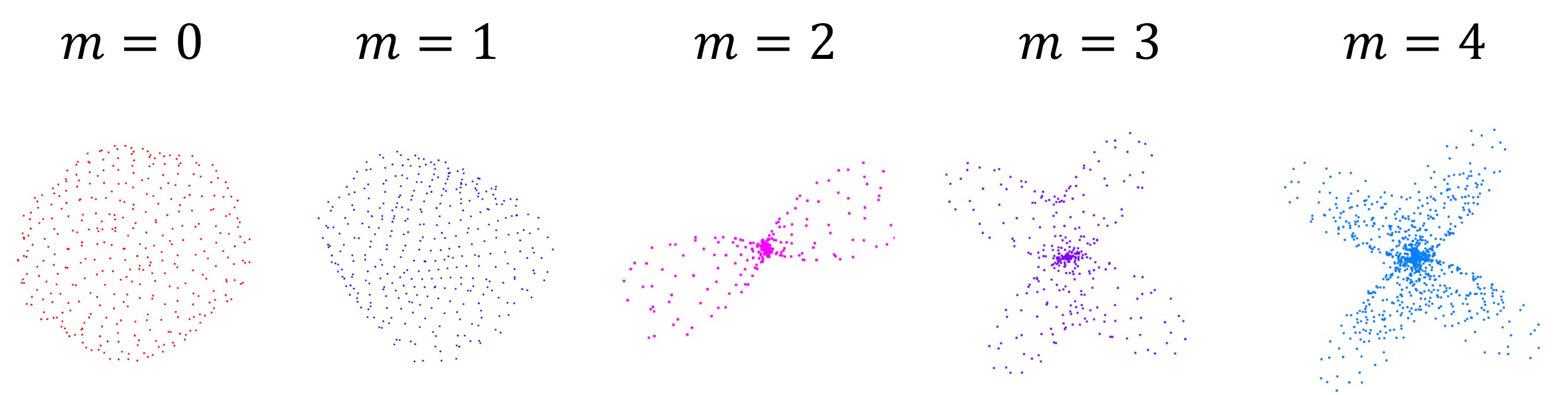}  
	\caption{
		Visualization results of the first five spherical spectrums
	}
	\label{fig:drinet}
\end{figure}
\subsection{Directional Sampling}
We use the Fibonacci spherical sampling strategy to generate a basic set of spherical unit vectors $\{\omega_l\}_{l=1}^{N_{\text{dir}}}$:\\
First, calculate the polar angle, using the cosine values evenly distributed over $[-1,1]$, map them to the polar range.
\begin{equation}
	\theta_i=\arccos\left(1-\frac{2\alpha_i}{N_{\text{dir}}}\right)\,,
\end{equation}
where $\alpha_i=i+0.5,\ i=0,1,2,\ldots,N_{\text{dir}}-1$.\\
Then calculate the azimuth, using the idea of the golden angle, so that the points are evenly spread out in the azimuth direction, i.e.,
\begin{equation}
	\phi_i=\alpha_i\times\phi_0,\quad \text{where}\ \phi_0=\pi(1+\sqrt{5}).
\end{equation}
Finally, using the conversion formula that utilizes standard spherical coordinates to Cartesian coordinates:
\begin{equation}
	x_i=\sin\theta_i\cos\phi_i,\quad y_i=\sin\theta_i\sin\phi_i,\quad z_i=\cos\theta_i\,.
\end{equation}
To get the final $\omega\in\mathbb{R}^{N_{\text{dir}}\times3}$.\\

\subsection{Computational Complexity}
There are $L$ directions and $M$ frequencies. For my method:\\
Each point and each direction vector need to calculate three times of multiplication and addition operation, and the total complexity is $\mathcal{O}(3NL)$. Each point calculates a complex index for each direction and accumulates it, and the complexity is $\mathcal{O}(NML)$, and the total complexity is
\begin{equation}
	\mathcal{O}\left(3NL\right)+\mathcal{O}\left(NML\right)=\mathcal{O}(NML)\,.
\end{equation}
Compared with spherical harmonic basis function expansion:\\
The spherical harmonic function $Y_l^m$ involves associated Legendre polynomials and complex exponential terms. Due to recursive calculation, if it is expanded to the maximum order $L_{\max}$, the computational complexity of each basis function at each point is $\mathcal{O}(N(L_{\max}+1)^2)$. At the same time, all points of each basis function need to be traversed and weighted, and the complexity is $\mathcal{O}(N(L_{\max}+1)^2)$. The total complexity is
\begin{equation}
	\mathcal{O}\left(N\left(L_{\max}+1\right)^2\right)+\mathcal{O}\left(N\left(L_{\max}+1\right)^2\right)=\mathcal{O}(NL_{\max}^2)\,.
\end{equation}
In practice, $M=32$, $L=60$, At this time, the total complexity of spherical harmonic expansion is
\begin{equation}
	\mathcal{O}\left(2\times61^2N\right)=\mathcal{O}(7442N)\,.
\end{equation}
The total complexity of DASFT is
\begin{equation}
	\mathcal{O}\left(3\times60N+32\times60N\right)=\mathcal{O}(2100N)\,.
\end{equation}
The time consumption is about $\frac{2100}{7442}\approx28.2\%$ of the spherical harmonic method.\\

At the same time, because DASFT is directly implemented by point product and complex exponential operation, it does not need pre-calculation, which has the advantage of parallelization compared with spherical harmonic basis function expansion which depends on recursive calculation. Block computing can be further used to reduce memory consumption.\\
We can divide the $N_{\text{dir}}$ direction into $K=\left\lceil\frac{N_{\text{dir}}}{\text{chunk\_size}}\right\rceil$ blocks. The group of unit vectors within each tile is $\{\omega_l\}_{l=1}^{K}$, where $l\in k\cdot \text{chunk\_size},k\cdot \text{chunk\_size}+1,\ldots,\min\left((k+1)\cdot \text{chunk\_size},N_{\text{dir}}-1\right)$. The energy spectrum is calculated by each piece and finally added.
\begin{equation}
	G(r)=\sum_{k=1}^K G_{\text{partial}}(r)=\frac{1}{N_{\text{dir}}}\sum_{k=1}^K\sum_{l=1}^{\text{chunk\_size}}E(r\cdot\omega_l)\,.
\end{equation}

\subsection{Error Analysis}
In the actual calculation, the spherical area fraction is approximated by uniformly sampling $N_{\text{dir}}$ directions $\{\omega_l\}_{l=1}^{N_{\text{dir}}}$, so we analyze the error. The Monte Carlo approximation of spherical area fraction $G(r) = \frac{1}{4\pi}\int_{S^2} E(r\cdot\omega)d\omega$ is:
\begin{equation}
	G(r) \approx G_{\text{true}}(r) = \frac{1}{N_{\text{dir}}}\sum_{l=1}^{N_{\text{dir}}} E(r\cdot\omega_l)\,,
\end{equation}
where $\{\omega_l\}$ is the direction of uniform sampling. According to Monte Carlo integration theory, the expectation of approximation error meets:
\begin{equation}
	\mathbb{E}\left[\left|G(r)-G_{\text{true}}(r)\right|\right] \leq \frac{\sigma_E}{\sqrt{N_{\text{dir}}}}\,,
\end{equation}
where $\sigma_E$ is the standard deviation of the energy spectrum $E(r\cdot\omega)$ on the sphere. Assuming that the maximum value of $E(r\cdot\omega)$ is $E_{\max}$, and the minimum value is $E_{\min}$, the upper bound of variance is:
\begin{equation}
	\sigma_E \leq \frac{E_{\max}-E_{\min}}{2}\,.
\end{equation}
So the error is:
\begin{equation}
	\left|\Delta G(r)\right| \leq \frac{E_{\max}-E_{\min}}{2\sqrt{N_{\text{dir}}}}\,.
\end{equation}
At the same time, we use layer normalization in the network to standardize each invariant $G(r)$:
\begin{equation}
	\hat{G}(r) = \frac{G(r)-\mu}{\sigma}\,.
\end{equation}
Where $\mu$ is the global mean and $\sigma$ is the standard deviation. So the normalized error is defined as:
\begin{equation}
	\Delta\hat{G}(r) = \hat{G}(r)-\hat{G}_{\text{true}}(r) = \frac{G(r)-\mu}{\sigma} - \frac{G_{\text{true}}(r)-\mu_{\text{true}}}{\sigma_{\text{true}}}\,.
\end{equation}
We assume that the normalized parameter is close to the true value, with $\Delta\mu=\mu-\mu_{\text{true}}$, $\Delta\sigma=\sigma-\sigma_{\text{true}}$, so:
\begin{equation}
	\Delta\hat{G}(r) \approx \frac{\Delta G(r)}{\sigma_{\text{true}}} - \frac{\Delta\mu}{\sigma_{\text{true}}} - \frac{\Delta\sigma(G_{\text{true}}(r)-\mu_{\text{true}})}{\sigma_{\text{true}}^2}\,.
\end{equation}
Due to the above assumptions (the influence of sampling error on the mean and standard deviation can be ignored), there are
\begin{equation}
	\Delta\hat{G}(r) \approx \frac{\Delta G(r)}{\sigma_{\text{true}}}\,.  
\end{equation}
The upper bound of the normalized error is obtained by substituting the upper bound of the original error:
\begin{equation}
	\left|\Delta\hat{G}(r)\right| \leq \frac{E_{\max}-E_{\min}}{2\sigma_{\text{true}}\sqrt{N_{\text{dir}}}}\,.
\end{equation}

To calculate the ratio error:
\begin{equation}
	\epsilon = \frac{\left|\Delta\hat{G}(r)\right|}{\left|\hat{G}_{\text{true}}(r)\right|}\,.
\end{equation}

Bring in actual data for calculation:
\begin{equation}
	\left|\Delta\hat{G}(r)\right| = \frac{1048576-0.487548}{2\times26573496\times\sqrt{36}} = 0.003288\,.
\end{equation}

Calculate average ratio error:
\begin{equation}
	\epsilon_{\text{avg}} = \frac{1}{N_r}\sum_{m=1}^{N_r}\frac{\left|\Delta\hat{G}(r)\right|}{\left|\hat{G}_{\text{true}}(r_m)\right|} = 0.928\%\,.
\end{equation}

When $N_{\text{dir}}=60$, $\epsilon_{\text{avg}}=0.366\%$ and when $N_{\text{dir}}=12$, $\epsilon_{\text{avg}}=7.821\%$.  
In addition, we compared the performance of the model under different $N_{\text{dir}}$ conditions.

\section{Hyperparameter Selection}
We determined the values of key hyperparameters through a series of experiments.

\begin{table}[b]
	\centering
	\begin{tabular}{lccc}
		\toprule
		\textbf{k} & \textbf{z/z} & \textbf{z/SO(3)} & \textbf{Params} \\
		\midrule
		\multicolumn{4}{c}{\textbf{DLP Method}} \\
		\midrule
		8 & 87.0 & 87.0 & 6.73M \\
		12 & 87.5 & 87.5 & 7.38M \\
		20 & \textbf{87.7} & \textbf{87.6} & 9.32M \\
		\midrule
		\multicolumn{4}{c}{\textbf{SAP Method}} \\
		\midrule
		8 & 86.5 & 86.3 & 6.34M \\
		12 & 87.1 & 87.0 & 6.42M \\
		20 & \underline{87.5} & \underline{87.5} & 6.60M \\
		\bottomrule
	\end{tabular}
	\caption{Performance comparison of DLP and SAP methods under different k values. The overall optimal results are shown in bold and the sub-optimal items are underlined.}
	\label{tab:dlp_sap_comparison}
\end{table}

\subsection{Local Neighbor Number k}
For the neighbor number k in local feature aggregation, we evaluated the performance of DLP and SAP methods at different k values on the ScanObjectNN dataset. As shown in Table~1, in the DLP method, as k increases, the model's performance under both z/z and z/SO(3) conditions shows an upward trend, achieving optimal performance at k=20 under the z/SO(3) condition, but the model parameters also increase accordingly. When using the SAP method, suboptimal performance is obtained at k=20 under the z/SO(3) condition. Considering the balance between computational efficiency and model performance, we set k=12 for the DLP method and k=20 for the SAP method in the final model.

\subsection{Global Direction Sampling Density $N_{\text{dir}}$}
For the spherical sampling number $N_{\text{dir}}$ in global direction perception, we evaluated the impact of different sampling densities on the model's rotation invariance. As shown in Table~2, when $N_{\text{dir}}$ is less than 24, the model's rotation invariance is significantly affected due to overly sparse sampling, leading to a noticeable drop in accuracy. Continuing to increase $N_{\text{dir}}$ beyond 36 yields negligible performance improvements. Therefore, to balance computational efficiency and model performance, we selected $N_{\text{dir}}$=36 as the final configuration, which ensures sufficient rotation invariance while avoiding unnecessary computational overhead.

\begin{table}[htbp]
	\centering
	\begin{tabular}{lc}
		\toprule
		$N_{\text{dir}}$ & \textbf{Acc. (\%)} \\
		\midrule
		6  & 89.9 \\
		12 & 90.7 \\
		24 & 91.2 \\
		36 & 91.4 \\
		48 & 91.3 \\
		60 & 91.4 \\
		72 & 91.3 \\
		\bottomrule
	\end{tabular}
	\caption{Results of different $N_{\text{dir}}$ (corresponding to sampling density) under the condition of z/SO(3).}
	\label{tab:dir_sampling}
\end{table}

\subsection{Frequency Range}
For the frequency range $[f_{\min}, f_{\max}]$ and sampling method, we conducted experiments as shown in Table 7. The results demonstrate high stability and negligible performance differences between linear and logarithmic sampling, as well as across different ranges. Therefore, we selected the empirical range of $[0, 12]$ with linear sampling for an efficient and stable configuration.

\begin{table}[htbp]
	\centering
	\begin{tabular}{ccc}
		\toprule
		\multirow{2}{*}{\textbf{Range} $[f_{\min}, f_{\max}]$} & \multicolumn{2}{c}{\textbf{Acc. (\%)}} \\
		\cmidrule(lr){2-3}
		& \textbf{Linear} & \textbf{Logarithmic} \\
		\midrule
		$[0, 12]$ & 91.4 & 91.4 \\
		$[0, 24]$ & 91.3 & 91.3 \\
		$[0, 36]$ & 91.4 & 91.2 \\
		\bottomrule
	\end{tabular}
	\caption{Performance comparison of different frequency ranges and sampling methods under z/SO(3) condition.}
	\label{tab:freq_range_comparison}
\end{table}

\subsection{Robustness to Hyperparameters}
A key property of our model is its robustness to the precise selection of its main hyperparameters: the directional sampling density $N_{\text{dir}}$ and the frequency range $[f_{\min}, f_{\max}]$. 
As shown in Table 6, model performance saturates at $N_{\text{dir}}=36$ and remains stable with further increases. 
Similarly, Table 7 shows that performance is consistently high and stable across different frequency ranges and sampling methods (linear vs. logarithmic).
This insensitivity is primarily due to the layer normalization applied to the generated invariants. This normalization effectively standardizes the feature space, making the model robust to these variations and significantly simplifying the tuning process.

\section{Limitation}

Although our proposed DiPVNet framework has achieved promising results, it is important to acknowledge several limitations. On one hand, while our atomic dot-product operators significantly reduce computational complexity compared to spherical harmonic expansions, the spherical response averaging process still requires considerable computational resources, particularly for large-scale point clouds. This may limit the potential for real-time applications in resource-constrained scenarios. On the other hand, our method assumes that the point cloud is relatively complete. When processing extremely sparse or heavily occluded point clouds, the model's performance may degrade, as directional information becomes insufficient for robust feature extraction.

\section{Computational Resources}

All experiments in this study were conducted on a high-performance computing server with the following specifications:

\begin{itemize}
	\item \textbf{GPU}: 1× NVIDIA H20-NVLink with 96GB memory
	\item \textbf{CPU}: 16 vCPU AMD EPYC 9K84 96-Core Processor
	\item \textbf{Memory}: 150GB RAM
\end{itemize}

\end{document}